\documentclass{article}
\usepackage{spconf,amsmath,graphicx}
\usepackage{amssymb}
\usepackage{float}
\usepackage[hidelinks]{hyperref}
\usepackage{bbm}
\usepackage{amsmath}
\usepackage[export]{adjustbox}
\usepackage{caption}
\usepackage{subfig}

\usepackage{pgfplots}

\def\SE{{K_\text{SE}}}
\def\SSE{{S_\text{SE}}}

\def\X{{\mathbf X}}
\def\E{{\mathbb E}}

\def\t{{\mathbf t}}
\def\y{{\mathbf y}}
\def\I{{\mathbf I}}
\def\R{{\mathbb R}}

\def\cF{{\mathcal F}}
\def\E{{\mathbb E}}
\def\ciF{{\mathcal F}^{-1}}
\usepackage{xcolor}

\title{Low-pass filtering as Bayesian inference}
\name{Cristóbal Valenzuela$^{\star}$ \qquad Felipe Tobar$^{\star}$$^{\dagger}$}
\address{$^{\star}$Departament of Mathematical Engineering, Universidad de Chile.\\
$^{\dagger}$Center for Mathematical Modeling, Universidad de Chile.}

\begin{document}
\ninept
\maketitle
\begin{abstract}
We propose a Bayesian nonparametric method for low-pass filtering that can naturally handle unevenly-sampled and noise-corrupted observations. The proposed model is constructed as a latent-factor model for time series, where the latent factors are Gaussian processes with non-overlapping spectra. With this construction, the low-pass version of the time series can be identified as the low-frequency latent component, and therefore it can be found by means of Bayesian inference. We show that the model admits exact training and can be implemented with minimal numerical approximations. Finally, the proposed model is validated against standard linear filters on synthetic and real-world time series. 
\end{abstract}
%


\section{Introduction}
Monitoring the spectral content of a time series is of critical importance in real-world applications across a wide range of scientific disciplines. This is because the concentration of energy at a specific range of frequencies might be indicative of mechanical faults \cite{Kwak2013FaultDO}, cardiac anomalies \cite{Tobar2017RecoveringLS}, astronomical discoveries \cite{Whitcomb:80,lerko}, and whale calls from submarine audio recordings \cite{whales} to name a few. 

The standard practice to isolate components within a specific frequency range from a time-series observation, referred to as \emph{filtering}, is to convolve the observations with an object called \emph{linear filter}. This convolution removes all frequencies that do not correspond to the desired frequency range, thus, \emph{filtering out} unimportant frequencies. The theoretical rationale behind this approach is supported by the application of the Convolution Theorem \cite{katznelson_2004} to power spectral densities (PSD): the PSD of a filtered time series corresponds to the PSD of the linear filter (user-designed) multiplied by the PSD of the observed time series (not controllable). This result allows for designing the linear filter so as to remove unwanted frequency components to then perform the numerical convolution. We refer to low-pass filtering when the range of frequencies to be removed are centred (symmetrically) around zero.

We identify two drawbacks of this standard approach to filtering. First, to perform the numerical convolution, the time series has to be uniformly sampled, that is, no missing observations or random acquisition times can be allowed. This is a rather stringent assumption, since in real-world applications missing data is commonplace due to mechanical or electrical failures and the sampling rate is given by the hardware. For instance, the observations of light curves in Astronomy are only available at some time instants due to climate conditions, orientation of the telescope and even the priority of the experiment within the agenda of the observatory. The second drawback of the convolution method is its implicit deterministic assumption: by computing a low-pass version of the time series as a moving average, we are accumulating observation noise without properly accounting for the dispersion that this might cause.

We aim to address these two drawbacks by formulating the low-pass filtering problem as a Bayesian inference one. We model the observed time series as a mixture of three latent components: one of low frequencies, one of high frequencies, and an observation noise component. Then, we find the low-frequency component through probabilistic inference: we place a prior distribution on each component and then, using  observations of the time series, we find the posterior distribution over the low-frequency component. In particular, we choose Gaussian processes \cite{Rasmussen:2005:GPM:1162254} priors over the components to leverage the expressiveness of the GP formulation while introducing minimal numerical approximations, therefore, the proposed method will be referred to as Gaussian process low-pass filter (GPLP).


\section{Background: Gaussian Processes}

\subsection{Spectral representation of Gaussian processes}

A Gaussian process (GP) \cite{Rasmussen:2005:GPM:1162254} over the input set $\X$ is a real-valued stochastic process $(f(x))_{x \in \X}$, such that for any finite subset of inputs $\{x_i\}_{i=1}^{n}\subset \X$, the random variables $\{f(x_i)\}_{i=1}^{n}$ are jointly Gaussian. Without loss of generality we choose $\X=\mathbb{R}^N$. In this sense, a GP defines a distribution over real-valued functions $f: \X\mapsto\R, x\rightarrow f(x)$, that is uniquely determined by its mean function $m(x) = \E(f(x))$, typically assumed to be zero, and its covariance {kernel} $K(x, x') = \text{cov}(f(x), f(x')), \ x,x' \in \X$. 

The covariance kernel summarises the dynamic behaviour of the GP and thus it is key when designing GP models. In practice, we can rely upon the Wiener-Khinchin theorem \cite{correlation_theory}, which states that an integrable function $K: \mathbb{R}^N \mapsto \mathbb{C}$ is the covariance function of a weakly-stationary mean-square-continuous stochastic process $f: \mathbb{R}^N \mapsto \R$ if and only if it admits the representation
\begin{equation}
	K(\tau) = \int_{\mathbb{R}^n}  e^{2\pi i \omega^{T} \tau} S(\omega)d\omega,
\end{equation}
where $S(\omega)$ is a non-negative bounded function on $\mathbb{R}^n$ and $i$ denotes the imaginary unit. Henceforth, given a  kernel $K$, we will refer to $S$ as their power spectral density given by the above theorem; where the PSD $S(\omega)$ is the Fourier transform of the covariance kernel $K(\tau)$. This result allows us to encode spectral properties directly in the covariance function by first designing the PSD to then calculate the kernel as the inverse Fourier transform of the so designed PSD function.

The relationship between GPs and spectral representations has acquired attention recently in the machine learning community. For instance, covariance functions can be constructed in the spectral domain in parametric \cite{wilson13,nips17_tobar} and nonparametric \cite{nips15_tobar,npr_15b} ways. Additionally, the harmonic structure of GPs has been exploited to develop computationally-efficient sparse GP models by using \emph{inducing variables} in the spectral domain \cite{lazaro2010sparse,lazaro2009,hensman2016variational}. More recently, GPs have also been considered to address the spectral estimation problem, in particular, for nonuniformly-sampled data and detection of periodicities \cite{hensman,protopapas,tobar_nips18}. An open challenge in the spectral treatment of GPs, is that learning frequency representations is prone to local optima, since one aims to approximate a periodogram; this has been partially addressed using Bayesian optimisation with derivatives \cite{bayesopt-sm} and derivative-free Monte Carlo methods \cite{ijcnn18}. In our case, however, training simply involves a standard square exponential kernel (presented next) and therefore optimisation is straightforward. 

\subsection{The square exponential case}

The \emph{de facto} covariance kernel for GP models is the square-exponential covariance denoted by
\begin{equation}
	\SE(x,x') = \sigma^2\exp\left(-\frac{1}{2l^2}||x-x'||^2\right),
	\label{eq:SE}
\end{equation}
where the parameter $\sigma^2$ denotes the marginal variance of the process (i.e., the magnitude) and $l$ denotes the lengthscale, that is, the range of correlation between values of the process: the larger the lengthscale the longer the range of temporal correlations. 

The popularity of the SE kernel stems form its properties \cite{Rasmussen:2005:GPM:1162254}. The paths generated by a GP with an SE kernel are (i) dense in the space of continuous functions, (ii) infinite-times differentiable a.e. and (iii) smooth, meaning that their power spectral density is concentrated around zero. In fact, due to the exponential form of the Fourier operator, the PSD of the SE kernel is also  SE and given by 
\begin{equation}
	\SSE(\xi) = \cF\{\SE\}(\xi) = \sigma^2\sqrt{2\pi l ^2}\exp(-2\pi^2l^2\xi^2 ), \label{eq:SSE}
\end{equation}
where the lengthscale of this PSD (spectral domain) is now inversely proportional to the lengthscale of the covariance (temporal domain); this has key advantages when using GPs for spectral estimation \cite{tobar_nips18,lazaro2009}. This can be understood intuitively: a process with long-range correlations has low frequency energy (smooth), whereas a kernel with short-length correlations necessarily has high frequency components (rough). 

The generative model proposed in the next section will represent observed signals as a GP with SE covariance function composed of a mixture of (non overlapping) low- and high-frequency components.

\section{A latent-component generative model for Bayesian filtering}

We propose the following generative model for a continuous-time (latent) signal $(f(t))_{t\in\R}$ as a mixture of two components of the form
\begin{align}
	f(t) &= f_{l}(t) + f_{h}(t),
	\label{eq:gen_mod}
\end{align}
where $f_{l}$ is a signal of low-frequency content and $f_{h}$ one of high-frequency content. 

\subsection[Assumptions over the spectral components]{Assumptions over the spectral components $f_l$ and $f_h$}
We model $f_{l}$ and $f_{h}$ as independent GPs with covariance kernels denoted respectively by $K_l$ and $K_h$, and, accordingly, power spectral densities $S_l$ and $S_h$. To discriminate between higher and lower frequencies, we impose the following restrictions over $S_l$ and $S_h$: 
\begin{enumerate}
	\item The support of $S_l$, denoted by $\text{supp}(S_l)$, is compact and centred around the origin, meaning that $f_l$ is a process of low-frequency content. 
	\item The supports of the PSDs of $f_l$ and $f_h$ are non overlapping, that is, $\text{supp}(S_l)\cap \text{supp}(S_h) = \emptyset$. This implies that each frequency present in the signal $f$ came, exclusively, from either $f_l$ or $f_h$.
	\item The sum of the component PSDs is a square-exponential kernel, that is, $S_l(\xi) + S_h(\xi) = \SSE(\xi)$ for some hyperparameters $\sigma^2$ and $l$ as in eq.~\eqref{eq:SE}.
\end{enumerate}
Notice that, as a consequence of the third restriction, the marginal distribution over the process $f$ is a Gaussian process with an SE kernel $\SE = K_l + K_h$ due to the linearity of the Fourier transform and the independence of $f_l$ and $f_h$. 

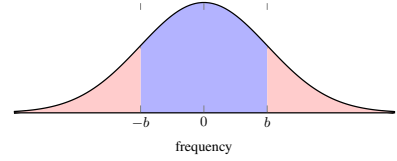
\begin{figure}
\centering 
\begin{tikzpicture}[thick, scale=0.6]
\begin{axis}[no markers, domain=-10:10, samples=100,
axis line style={draw opacity=0},
xlabel=frequency,
height=4cm, width=10cm, 
xtick={-1, 0, 1}, xticklabels={$-b$, $0$, $b$}, ytick=\empty,
enlargelimits=false, clip=false, axis on top]
\addplot [fill=none, draw=black, style={ultra thick}, domain=-3:3] {gauss(0,1)} \closedcycle;
\addplot [fill=blue!30, draw=none, domain=-1:1] {gauss(0,1)} \closedcycle;
\addplot [fill=red!20, draw=none, domain=-3:-1] {gauss(0,1)} \closedcycle;
\addplot [fill=red!20, draw=none, domain=1:3] {gauss(0,1)} \closedcycle;
\end{axis}

\end{tikzpicture}
\caption{Power spectral densities of the proposed model: The region contained inside the black line is the PSD of the process $f$, $\SSE$, whereas the regions in blue and red denote the PSDs of the low-frequency ($S_l$) and high-frequency ($S_h$) content respectively. Frequency $b$ is the maximum frequency in the support of $S_l$ and the minimum  in the support of $S_h$ (positive part).}
\label{fig:SlSh}
\end{figure}

Figure \ref{fig:SlSh} illustrates the PSDs of the components of high and low frequency. We have denoted by $b$ the interface between the zones of low and high frequency, meaning that $b$ is the highest frequency of the low-frequency signal and well as the lowest frequency of the high-frequency signal. Consequently, the bandwidth of the low-frequency part is $2b$. 
\vspace{1em}

\pgfmathdeclarefunction{gauss}{2}{\pgfmathparse{1/(#2*sqrt(2*pi))*exp(-((x-#1)^2)/(2*#2^2))}%
}

\subsection{Likelihood and model fitting}
\label{sec:fitting}
Assuming an independent sequence of Gaussian observation noise, the observations $(y(t))_{t\in\R}$ are then defined as
\begin{equation}
	y(t) = f(t) + \eta(t),\ \eta(t)\sim\mathcal{N}(0,\sigma_\eta^2),
	\label{eq:likelihood}
\end{equation}
Combining the observation model defined in eq.~\eqref{eq:likelihood} with the GP-prior assumed for the spectral components, the marginal likelihood of the proposed model is Gaussian and therefore its hyperparameters can be obtained through minimisation of the negative log-likelihood (NLL). Notice that despite the elaborate frequency-wise construction of the latent process $f$ through the non-overlapping spectra of the components $f_l$ and $f_h$, the covariance kernel of $f$ is square-exponential, thus allowing for straightforward model learning. Specifically, the NLL of the model is given by
\begin{align}
\text{NLL}(\y|\t) = \log(2\pi |\Sigma_\y|) + \frac{1}{2}\y^\top\Sigma_{\y}^{-1}\y,
\label{eq:NLL}
\end{align}
where $\y=[y_1,\ldots,y_N]$ are the (noise corrupted and possibly missing) observations acquired at time instants $\t=[t_1,t_2,\ldots,t_N]$, and $\Sigma_\y$ is the covariance matrix of $\y$ defined by  
\begin{equation}
	\Sigma_\y = \SE(\t,\t) + \sigma_\eta^2\I,
	\label{eq:Sigmaobs}
\end{equation}
therefore, the hyperparameters are those of the $\SE$ kernel and the noise variance $\sigma_\eta^2$. 

Finally, observe that the strict non-overlapping property of the components $f_l$ and $f_h$ is not problematic for training, in fact, the cutoff frequency $b$ does not even appear for model training.

\section{Filtering as posterior inference}


Denote by $b$ the required cut-off frequency of the low-pass filtering problem. Using the proposed model, we can assume that this cutoff frequency $b$ is equal to the limit between the low- and high-frequency components. In this context, low-pass filtering problem is equivalent to performing inference over the low-frequency component $f_l$ conditional to observations of the time series. Due to the assumptions made on the signal we refer to this approach as GP low-pass filter (GPLP).

Denoting the observations by $\y\in\mathbb{R}^n$, GPLP addresses low-pass filtering by computing the posterior distribution $p(f_l|\y)$. Due to the self-conjugacy of the Gaussian distribution and its closure under additivity, this posterior is also a GP, with  mean and covariance given by 
\begin{align}
m_{f_l\vert\y} & = \Sigma_{f_l,\y}\Sigma^{-1}_{\y}\y
\label{eq:post_mean}\\
K_{f_l\vert\y} & = K_{f_l} -  \Sigma_{f_l,\y}\Sigma^{-1}_{\y}\Sigma_{f_l,\y}^\top, \label{eq:post_var}
\end{align}
where we have assumed zero mean for $f_l$ and $f_h$ (and therefore of $y$), $\Sigma_{\y}$ is the covariance of the observations defined in eq.~\eqref{eq:Sigmaobs}, $\Sigma_{f_l,\y}$ denotes the covariance between $f_l$ and $\y$, and $K_{l}$ is the kernel of $f_l$. 

Let us also note that the cross covariance $\Sigma_{f_l,\y}$ and the kernel $K_{l}$ share the same expression. Denoting the covariance between the low-frequency process $f_l$ at time $t$ and the observation $y$ at time $t'$ by $\Sigma_{f_l,\y}(t,t')$, we obtain
\begin{align}
	\Sigma_{f_l,\y}(t,t') &= \E[f_l(t)(f_l(t')+f_h(t')+\eta(t'))]\nonumber\\
					&= \E[f_l(t)f_l(t')]\nonumber\\
					&= K_l(t,t'),\nonumber
\end{align}
since the processes $f_l$, $f_h$ and $\eta$ are independent Gaussian processes. 

Therefore, the only critical quantity required to compute eqs.~\eqref{eq:post_mean}-\eqref{eq:post_var} is kernel $K_{l}$. Following the model proposed in eq.~\eqref{eq:gen_mod} and its assumptions, the PSD of $f_l$, denoted by $S_l$, can be obtained by multiplying the PSD of $f$ with a rectangular function of width $2b$, that is, 
\begin{equation}
 	S_l(\xi) = \SSE(\xi)\text{rect}\left(\frac{\xi}{2b}\right),
 \end{equation} 
 where we used the convention that $\text{rect}(\xi)$ is equal to one for $|\xi|<1/2$ and 0 elsewhere. As a consequence, the kernel $K_l$ can be calculated using the convolution theorem: ($\star$ is the convolution operator)
\begin{align}
 	K_l(t) &= \ciF(S_l(\xi))\\
 		&= \ciF\left(\SSE(\xi)\text{rect}\left(\frac{\xi}{2b}\right)\right)\nonumber\\
 		&= \ciF\left(\SSE(\xi)\right)\star \ciF\left(\text{rect}\left(\frac{\xi}{2b}\right)\right)\nonumber\\
		&= \SE(t) \star  \text{sinc} \ (2bt) \cdot2b\nonumber\\
	&= 2b \cdot \int \sigma^2 \exp\left(-\frac{1}{2l^2}(t- \tau)^2\right) \frac{\sin(2\pi b\tau)}{2\pi b\tau} d\tau\nonumber\\
	&=  \sigma^2e^{-\frac{1}{2l^2}t^2}  \Re\left(\text{erf}\left(\sqrt{2}bl\pi-i\tfrac{t}{\sqrt{2}l}\right)\right),\nonumber
\end{align}
where $\text{erf}(t)$ denotes the error function given by 
\begin{equation}
	\text{erf}(t) = \frac{1}{\sqrt{\pi}}\int_{-t}^{t}e^{-x^2}dx,
\end{equation}
Using Taylor expansions, the error function can be calculated up to an arbitrary degree of accuracy \cite{CHEVILLARD201272}.


\section{Simulations }
The proposed model for Bayesian low-pass filtering using GPs, termed GPLP, is next validated using synthetic and real-world data. Our experimental validation aims to show that GPLP (i) successfully recovers low-frequency data from missing and noisy observation, (ii) provides accurate point-estimates with respect to the benchmarks, and (iii) places meaningful error bars. Our benchmarks include ground-truth signals and the Butterworth filter.

\subsection{A synthetic time series with line spectra}
\label{sec:evenly}

We considered the line-spectra time series given by 
\begin{equation}
	f(t) = \sum_{\omega_i \in F_{\text{low}}}\cos(2\pi \omega_it) + \sum_{\omega_j \in F_{\text{high}}} \cos(2\pi \omega_jt)
	\label{eq:line_spectra}	
\end{equation}
were the sets $F_{\text{low}}$ and $ F_{\text{high}}$ are such that $\forall \omega_i \in F_{\text{low}}, \forall \omega_j \in F_{\text{high}}: \omega_{i}<\omega_j$. Simply put, $F_{\text{low}}$ is a set of low frequencies and $F_{\text{high}}$ a set of high frequencies---all these frequencies are in Hertz (Hz). Signals constructed in this way have sparse PSDs meaning that only a finite number of frequencies convey all the signal energy or information.

\begin{figure}[t]
	\centering
	\includegraphics[width=1.0\linewidth]{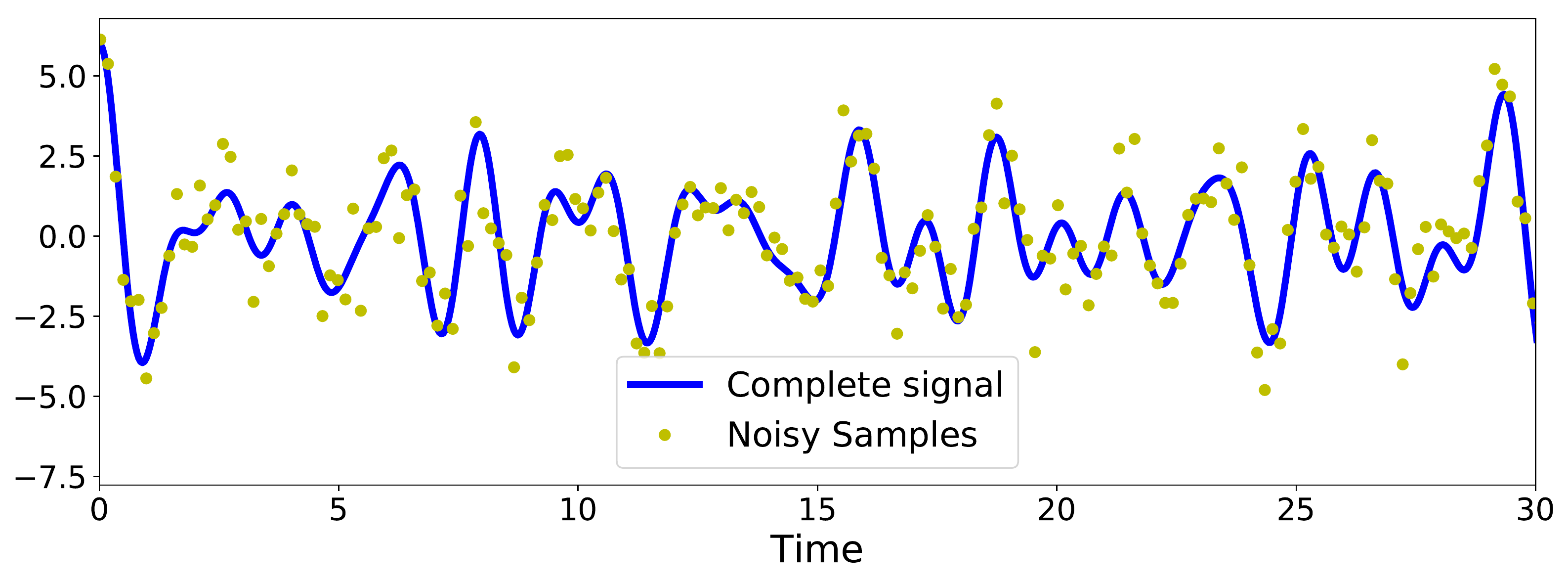}
	\caption{The latent signal (blue) consists of 6 cosines as shown in eq.~\eqref{eq:line_spectra} and the observations are shown in yellow.}
	\label{fig:data1}
\end{figure}

We chose $F_{\text{low}} = \{0.31,0.38,0.48\}$ and $F_{\text{high}} = \{0.51,$ $0.64,0.75\}$  and simulated a path of $f(t)$, as defined in eq.~\eqref{eq:line_spectra} for 5000 evenly-spaced time indices in $t\in[-100,100]$. The observation time-series $\y$ consisted only in a 25\% of the signal (again, evenly spaced) all of which were corrupted by Gaussian noise of std.~dev.~$\sigma_\eta=1.0$. Fig.~\ref{fig:data1} shows the latent signal and the observation considered for this experiment. 

\begin{figure}[t]
	\centering
	\includegraphics[width=1\linewidth]{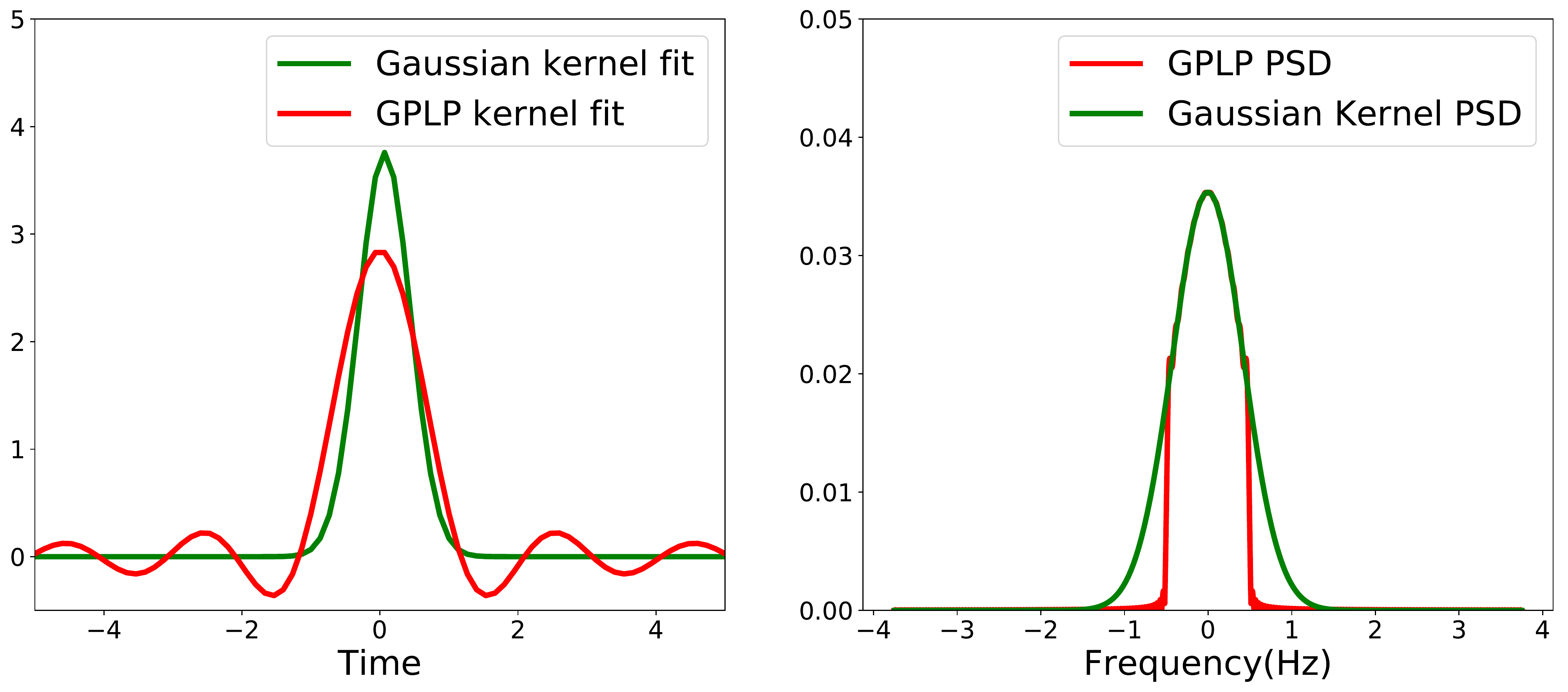}
	\caption{Left: Learnt Gaussian kernel shown in green and the low-frequency kernel in red. Right: Learnt Gaussian PSD shown in green and low-frequency PSD in red. }
	\label{fig:kernel1}
\end{figure}

We implemented the proposed GPLP to recover the low-frequency content of the original (latent) signal $f$ only using the observations $\y$. We first trained the generative model as explained in Sec.~\ref{sec:fitting} to find the hyper parameters $l$, $\sigma^2$ and $\sigma^2_\eta$. We then chose the cutoff frequency to be $b=0.495$Hz. We then computed the low-frequency covariance function to calculate the moments of the posterior distribution $p(f_h|\y)$. Fig.~\ref{fig:kernel1} shows the learnt kernels and their corresponding PSDs. Notice how, just as illustrated in Fig.~\ref{fig:SlSh}, the spectral densities of the latent low-frequency component is band-limited, supported only on $[-b,b]$ and tightly bounded by the (unfiltered) time series. 

\begin{figure}[t]
	\centering
	\includegraphics[width=1\linewidth]{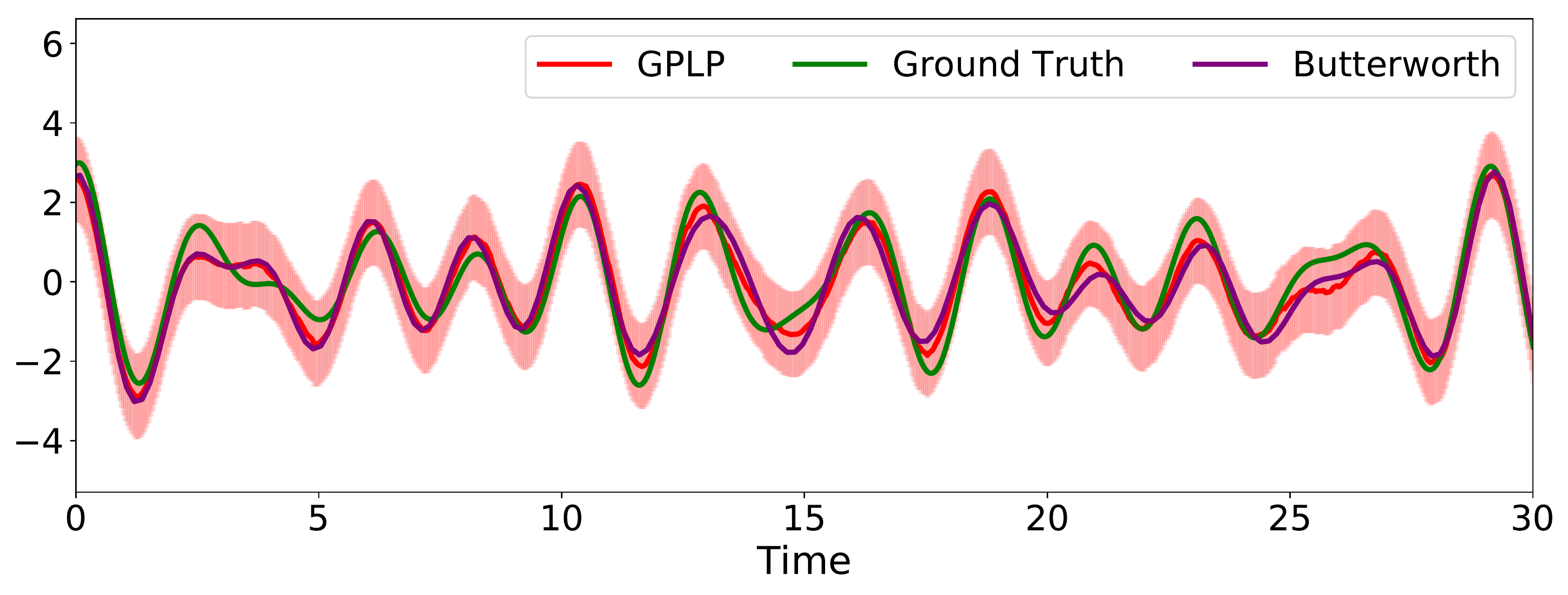}
	\caption{Inference over the low-frequency component: posterior mean of the proposed GPLP (red), Butterworth low-pass filter (purple) and ground truth signal (green).}
	\label{fig:post_mean}
\end{figure}

Fig.~\ref{fig:post_mean} shows the GPLP estimate compared against the  ground truth and a low-pass version of the data using a Butterworth low-pass filter of order $10$, with the same cutoff frequency; this filter is a standard in linear filtering. GPLP obtained a mean-squared error of $0.16$ while the Butterworth low-pass filter gave a mean-squared error of $0.26$, in addition to this marginal difference in performance, notice that GPLP provided accurate 95\% error bars. 

\begin{figure}[t]
	\includegraphics[width=1\linewidth, left]{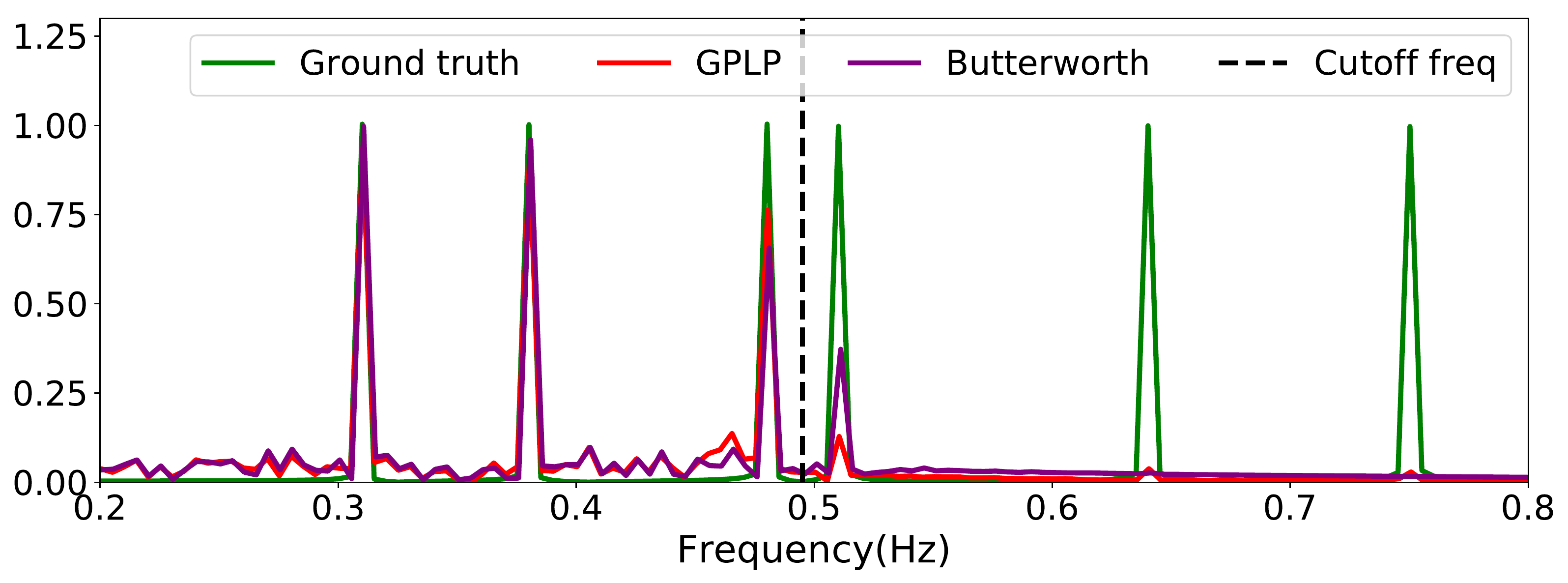}
	\caption{FFT for full signal (green) and low-pass estimates: proposed GPLP (red) and Butterworth (purple). The cutoff frequency is shown in  by a vertical dashed line. }
	\label{fig:FFT1}
\end{figure}

To further validate the ability of the proposed GPLP to filter out low-frequency spectral content, Fig.~\ref{fig:FFT1}  shows the Fast Fourier Transform (FFT) of the low-pass versions of GPLP, Buttwerworth and the original signal. Notice from that GPLP successfully recovered the first three spectral components and rejected the higher ones. 

\

\subsection{Low-pass filtering of unevenly-sampled observations}

A critical downside of the standard filtering techniques is that most of them require the data to be evenly spaced. Here is where the proposed model excels: as our method is based on an infinite-dimensional prior over continuous-time signals, missing observations are naturally handled by integrating out missing values. 

\begin{figure}[t]
	\includegraphics[width = 1.0\linewidth]{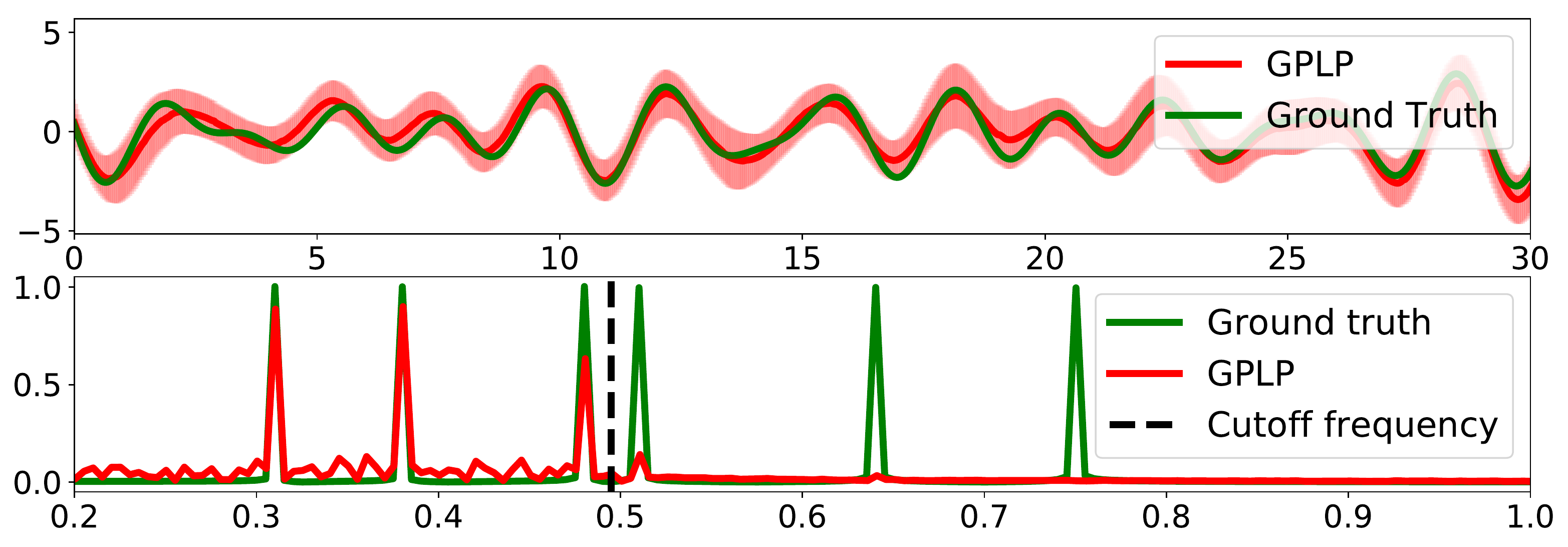}
	\caption{Above:Low-pass filtering of unevenly-sampled observations. Proposed GPLP shown in red (mean and 95\% error bars).
	Below: Same result, but shown in the frequency domain.}
	\label{fig:unevenly1}
	\centering
\end{figure}

We replicated the exact same setting  as in Sec.~\ref{sec:evenly} but considered \textbf{randomly-chosen} observations (Again, just $25 \%$ of the total number of points). Fig.~\ref{fig:unevenly1} shows the posterior mean over the low-frequency component together with the 95\% confidence interval and the ground truth, as well as the result in the frequency domain. The MSE of the GPLP estimate was $0.23$, thus improving over Butterworth using evenly spaced data.

\subsection{Filtering a heart-rate time series}

We considered two 1800-sample heart-rate signals\footnote{\url{http://ecg.mit.edu/time-series/}}. One corresponding to a healthy subject and an unhealthy subject. The anomally can be detected from the heart-rate signal by looking at the energy contained below $0.05$ (Hz): if most of the energy is below this threshold, the subject is likely to suffer from congestive heart failure \cite{Glass_Hunter_McCulloch_1991}. 

The aim of this experiment was to use GPLP to quantify the portion of energy below $0.05$, as this reveals whether the signal corresponds to a healthy or unhealthy subject. We implemented GPLP on both signals with a cutoff frequency of $0.05 (Hz)$ and we found that the healthy signal has a 77\% of its energy below $0.05 (Hz)$ and that the unhealthy signal has a 97\% of its energy below $0.05 (Hz)$. Therefore, the GPLP method can discriminate between healthy and unhealthy subjects from the heart-rate signals. 

\section{Discussion}

We  have proposed a Bayesian approach to low-pass filtering. The method is based on a latent-component generative model for time series, where the components are Gaussian processes with non-overlapping spectra. With this model, finding the low-pass version of a signal can be addressed from a Bayesian inference point of view. The main contribution over existing low-pass filters in the linear filter literature is that the proposed model offers an account of its own uncertainty and can naturally handle missing or noisy observations. 

The proposed method has been validated empirically using synthetic and real-world data, where we have shown its ability to recover unbiased estimates of the true low-frequency signals (both in the evenly- and unevenly-sampled cases) and performed accurately with respect to its classical counterpart: the Butterworth filter.



\section{Acknowledgements}
This work was funded by projects Fondecyt-Iniciación \#11171165 and Conicyt-PIA \#AFB170001 Center for Mathematical Modeling. 

\newpage

\bibliographystyle{IEEEbib}
\bibliography{biblio}

\end{document}